\documentclass{article}

\PassOptionsToPackage{square, numbers}{natbib}

     \usepackage[final]{tackling_climate_workshop_style}

\usepackage[utf8]{inputenc} 
\usepackage[T1]{fontenc}    
\usepackage{hyperref}       
\usepackage{url}            
\usepackage{booktabs}       
\usepackage{amsfonts}       
\usepackage{nicefrac}       
\usepackage{microtype}      
\usepackage{graphicx}
\usepackage{bbold}
\usepackage{subcaption}

\title{Image-based Early Detection System for Wildfires}

\author{%
    Omkar Ranadive\thanks{Equal Contribution} \\
    Alchera X\\
    \texttt{omkar.ranadive@alcheralabs.com} \\
    \And
    Jisu Kim\textsuperscript{*}\\
    Alchera \\
    \texttt{js.kim@alcherainc.com} \\
    \AND
    Serin Lee\textsuperscript{*}\\
    Alchera X \\
    \texttt{serin.lee@alcheralabs.com} \\
    \And 
    Youngseo Cha \\
    Alchera \\
    \texttt{ys.cha@alcherainc.com} \\ 
    \And 
    Heechan Park \\
    Alchera \\ 
    \texttt{hc.park@alcherainc.com} \\ 
    \AND
    Minkook Cho \\ 
    Alchera \\ 
    \texttt{mk.cho@alcherainc.com} \\ 
    \And
    Young K. Hwang\thanks{Corresponding author} \\
    Alchera \\
    \texttt{yk.hwang@alcherainc.com} 
}

\begin{document}

\maketitle

\begin{abstract}
Wildfires are a disastrous phenomenon which cause damage to land, loss of property, air pollution, and even loss of human life. Due to the warmer and drier conditions created by climate change, more severe and uncontrollable wildfires are expected to occur in the coming years. This could lead to a global wildfire crisis and have dire consequences on our planet. Hence, it has become imperative to use technology to help prevent the spread of wildfires. One way to prevent the spread of wildfires before they become too large is to perform early detection i.e, detecting the smoke before the actual fire starts. In this paper, we present our Wildfire Detection and Alert System which use machine learning to detect wildfire smoke with a high degree of accuracy and can send immediate alerts to users. Our technology is currently being used in the USA to monitor data coming in from hundreds of cameras daily. We show that our system has a high true detection rate and a low false detection rate. Our performance evaluation study also shows that on an average our system detects wildfire smoke faster than an actual person. 
\end{abstract}

\section{Introduction}

Every year, wildfires burn down millions of acres of land, damage thousands of structures and they also cause severe air pollution. Wildfires are disastrous to both human communities and wildlife. They cause billions of dollars in damages every year. In 2021 alone, there were over 58 thousand fires in the USA which burned down 7.13 million acres of land. Due to the global warming caused by climate change \cite{lindsey2020climate, philip2021rapid}, warmer and drier conditions are being created. Warmer temperatures cause greater evaporation which leads to drier soils and vegetation which makes them more susceptible to burning and can lead to severe wildfires. Studies have shown that in the coming decades, there is going to be a global increase in large and severe wildfires due to the adverse effects of climate change \cite{wuebbles2017climate, sullivan2022spreading, westerling2016increasing}. Wildfires are already extremely difficult to control and mitigate and if their severity keeps worsening over the years like the studies have shown then it would have dire consequences on our planet. 

Over the years, technology has been used to analyze different aspects of wildfires like - Fire Spread Simulators \cite{finney1998farsite, ramirez2011new} and Detecting Wildfires from Satellite Data \cite{toan2019deep, kyzirakos2014wildfire}. There are also many different fire management tools available online. While these technological tools are helpful in analyzing and mitigating the spread of wildfires, they don't help much in early detection and prevention of wildfires. 

\begin{figure}
  \centering
  \includegraphics[width=\linewidth]{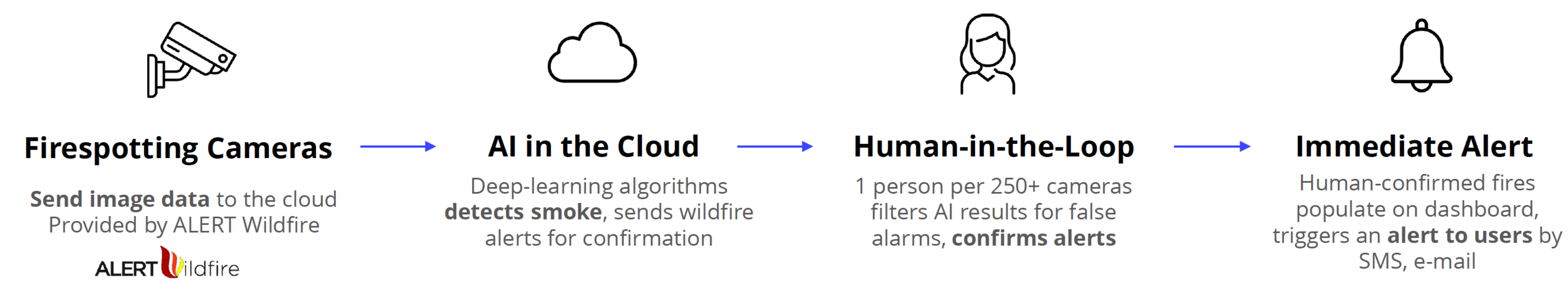}
  \caption{Overview of our Wildfire Detection and Alert System}
  \label{system}
\end{figure}

Cameras providing real-time image feed have been used for quite some time by human operators to monitor and identify wildfires. However, it is not possible for humans to monitor such data accurately around the clock. Machine learning (ML) algorithms are much more efficient for such real-time image based data. Such ML algorithms can perform early detection of wildfires i.e, detecting the smoke before the actual fire starts \cite{aslan2019early}. While there are many early detection products in the market, many of them require specialized hardware, have limited coverage, and have a high false detection rate. We present our Wildfire Detection System in this paper which has a high true detection rate, low false detection rate and which only uses real-time camera data to perform early detection of the wildfire smoke (Fig \ref{system}). The human operator now only has to confirm the detection alert instead of monitoring the cameras all the time. Once the alert is confirmed, authorities can reach the scene and prevent the wildfire smoke from turning into a large, uncontrollable wildfire. Our system is currently being actively used in the USA and monitors near real-time data from hundreds of cameras daily.

\section{Methodology}

\subsection{Data}

There are no good publicly available labeled datasets which can be used for early detection of wildfires. So we gathered raw image data from Alert Wildfire Cameras \cite{alertwild}.  Alert Wildfire is an association which provides open access to fire spotting cameras placed in different parts of the USA. These cameras are widely used by firefighters and first responders. We gathered data from 400 different cameras. These cameras cover many different terrains and the data was collected every month from Jun'21 - Mar'22. Therefore, the dataset is rich and diverse. The final dataset consists of 90,000 images. We then labeled the images with bounding boxes and categorized the images as wildire and non-wildfire images. More details on the dataset can be found in Appendix \ref{dataset}.

\subsection{Model Architecture}
\begin{figure}
  \centering
  \includegraphics[width=0.9\linewidth]{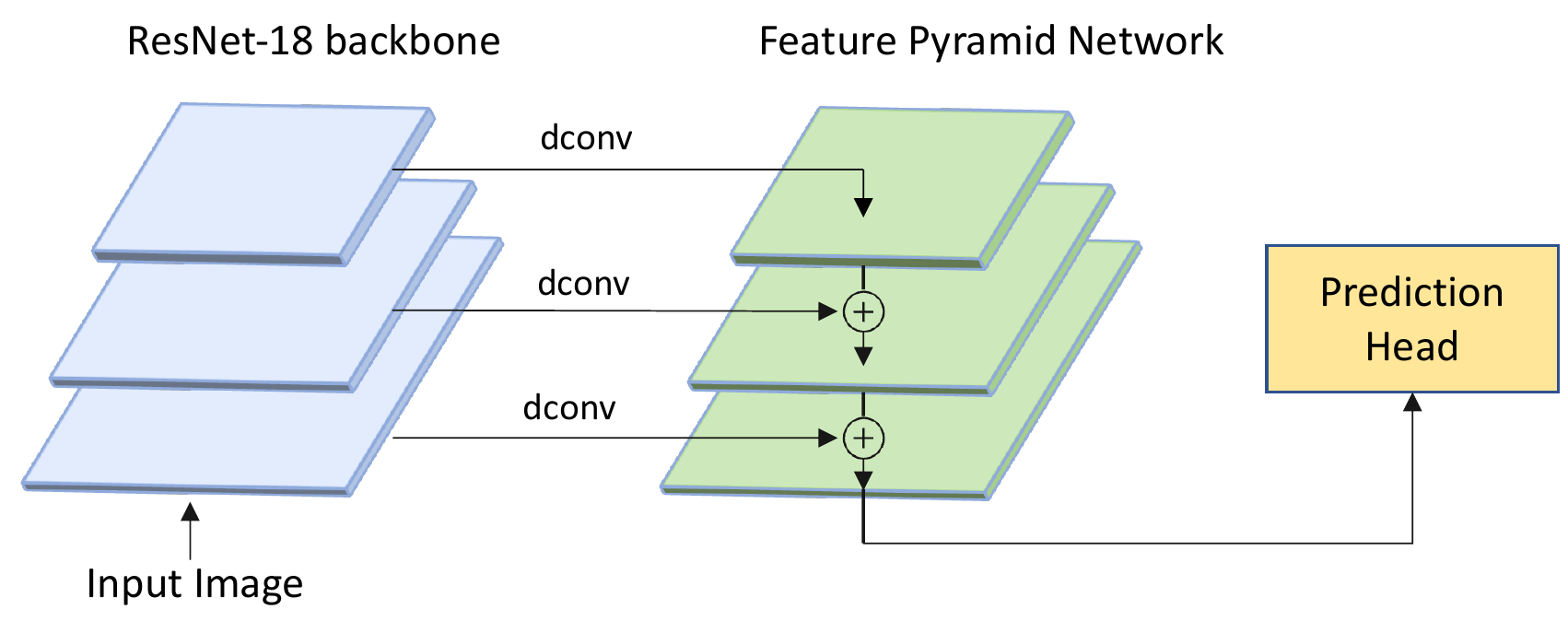}
  \caption{Model Architecture of our Wildfire Detection System}
  \label{architecture}
\end{figure}

The main challenge with smoke detection is that smoke can be of varying scales. Therefore, our model architecture is specifically designed to extract relevant information at different scales. To do this, we use an architecture similar to the one used in FCOS \cite{tian2019fcos}. An overview of our architecture can be seen in Fig \ref{architecture}. As our detection system gets near real-time data from multiple cameras, we needed a backbone network which is relatively lightweight and fast. That's why we chose ResNet-18 \cite{he2016deep} as our backbone network. 

After the input image is passed through our backbone ResNet-18, the Feature Pyramid Network (FPN) \cite{lin2017feature} is then used to extract information from the different layers of ResNet-18. Inside the FPN, we apply deformable convolution operation (dconv) \cite{dai2017deformable} to the information extracted from each layer. Deformable convolutions have offsets which are learned based on the target task. Hence, as opposed to the fixed receptive field of a normal convolution operation, deformable convolutions allow us to have an adaptive receptive field which is more useful to detect smoke at various scales. 

After the deformable convolution operation, the information at every level of the pyramid is added together. Then a set of 1x1 convolution operations are applied in the prediction head to get a classification (wildfire vs non-wildfire) and the bounding box coordinates. The bounding box coordinates are obtained using regression as a 4D vector $t^{*}$, similar to how they are obtained in FCOS \cite{tian2019fcos}.

For training, we use a loss function similar to that of FCOS \cite{tian2019fcos}. It is defined as follows: 

\begin{equation}
    loss = \frac{1}{N_{pos}}(\sum_{x, y}L_{cls}(p_{x, y}, c_{x, y}) + \sum_{x, y}\mathbb{1}_{\{c_{x, y} = 1\}}L_{reg}(t_{x, y}, t^{*}_{x, y}) +  \sum_{x, y}\mathbb{1}_{\{c_{x, y} = 1\}}L_{cen}(t^{*}_{x, y})). 
\end{equation}

where $N_{pos}$ is the number of positive samples, $L_{cls}$ is focal loss \cite{lin2017focal}, $p_{x, y}$ is the predicted score and $c_{x, y}$ is the grouth truth label, $L_{reg}$ is IOU loss \cite{yu2016unitbox} where $t_{x, y}$ is the vector of ground truth bounding box coordinates and $t^{*}_{x, y}$ is the vector of the predicted coordinates, and $L_{cen}$ is centerness loss \cite{tian2019fcos}. $L_{reg}$ and $L_{cen}$ are only calculated for positive samples. A sample is considered positive if  $c_{x, y} = 1$ and if it gets selected using our training sample selection process. 

\begin{figure}
  \centering
  \includegraphics[width=\linewidth]{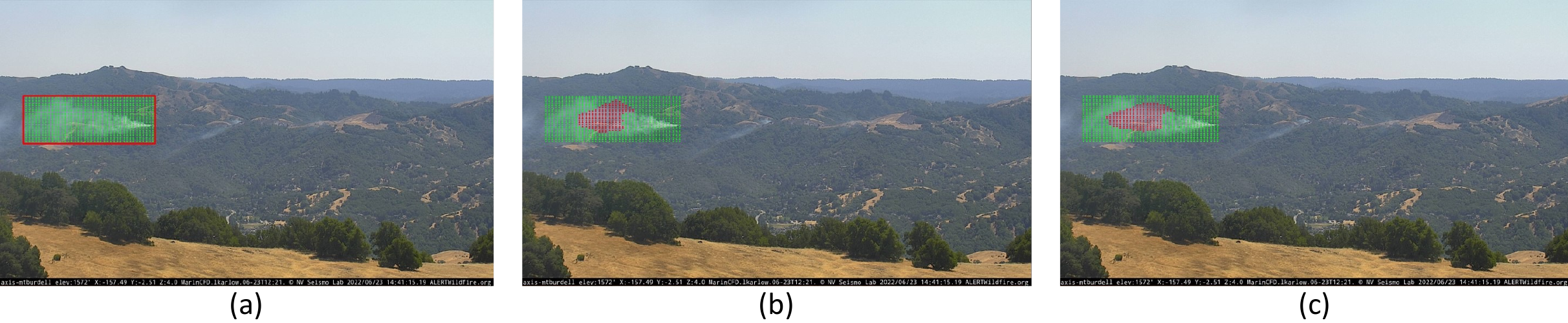}
  \caption{Adaptive Training Sample Selection Process. (a) All points inside the bounding box are considered as candidate samples. (b) Positive samples selected using original ATSS  (c) Positive samples selected using our modified ATSS}
  \label{atss}
\end{figure}

\subsection{Modified Adaptive Training Sample Selection}
In a recent paper \cite{zhang2020bridging}, it was shown that the performance of object detectors is dependent on how the positive and negative samples are selected. The paper \cite{zhang2020bridging} also proposes an adaptive training sample selection (ATSS) process which selects these samples based on statistical characteristics of the object. In our sample selection process, we use a modified version of ATSS. 

The process begins by considering all points inside the ground truth bounding box as candidate samples (Fig \ref{atss}a). Then IOU is calculated between the bounding boxes predicted by the candidate samples and the ground truth bounding box. Then a final sample score is calculated for every candidate point as follows: 
\begin{equation}
    Score_{x, y} = (IOU)_{x,y} * p_{x, y}
\end{equation}
Where $(IOU)_{x, y}$ is the IOU value for the sample point and $p_{x, y}$ is the confidence value, i.e, output from prediction head of the model for that point. Finally, threshold is calculated as the mean + standard deviation of all scores. Then samples greater than the threshold value are selected as positive samples. The remaining samples inside the ground truth bounding box are not considered as negative samples but instead ignored during the loss calculation. Fig \ref{atss}b and Fig \ref{atss}c show the positive samples selected by original ATSS and our modified ATSS. Our version is able to select a greater number of relevant samples for training.

\begin{figure}
  \centering
  \includegraphics[width=\linewidth]{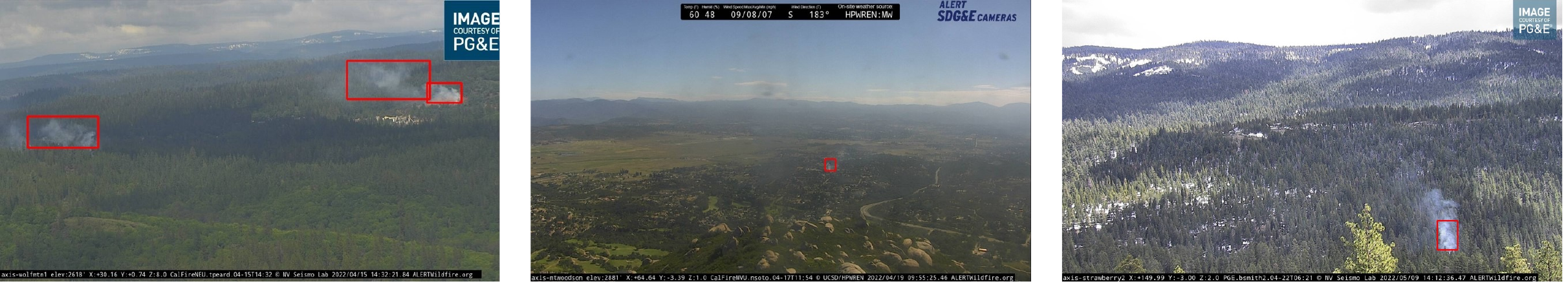}
  \caption{Examples of wildfire smoke detected by our model}
  \label{res}
\end{figure}

\begin{table}
  \centering
  \caption{Performance on validation set} 
  \begin{tabular}{lllll | llll}
  \toprule
    & TP & TN & FP & FN & Acc & Pre & TPR & FPR \\
  \midrule 
    No. of images    & 10690  & 27005 & 2205 & 1234 & 0.916 & 0.829  & 0.896 & 0.075 \\
  \bottomrule 
  \end{tabular}
  \label{perf}
\end{table}

\begin{table}[!hbt]
  \caption{Evaluation Study from Jun'22 - Jul'22}
  \label{eval}
  \centering
  \begin{tabular}{l | c}
    \toprule
    Time taken to detect wildfire smoke & \% of smoke events detected (cumulative) \\
    \midrule
     Within 60 seconds from the start of smoke  & 60.5\% \\
     Within 3 minutes from the start of smoke & 76.8\% \\
     Within 5 minutes from the start of smoke & 86.9\% \\
     Over 5 minutes from the start of smoke & 100\% \\
    \bottomrule
  \end{tabular}
\end{table}

\section{Results}
As seen in Fig \ref{res}, our model is capable of detecting multiple wildfire smoke events in the same image and also visually hard to detect smoke events. Table \ref{perf} shows the performance of our model on the validation set. It can be seen that the model has a high number of True Positives (TP) and True Negatives (TN) and a low number of False Positives (FP) and False Negatives (FN). The model has an accuracy (acc) of 91.6\%, precision of 82.9\%, a true positive rate (TPR) / recall of 89.6\% and a false positive rate (FPR) of 7.5\%. More instances of detection results can be found in Appendix \ref{detRes}. 

We also did an evaluation study of our model in the real-world over two months from Jun'22 - Jul'22. A total of 869 wildfire smoke events occurred during this period. In Table \ref{eval}, we show the time taken by our model to detect these smoke events. Our model was able to detect 60.5\% of the total events within just 60 seconds from the start of smoke. A total of 86.9\% of the events were detected within 5 minutes from the start of the smoke and all 869 events (100\%) were detected after 5+ minutes. We also compared the average detection time of the model with the average detection time from a human operator. The model on an average, detected the events 2 minutes, 57 seconds faster. These results show that our model is capable of fast and accurate detection. Details on challenges and future work can be found in Appendix \ref{cfwork}. 

\section{Conclusion}
In this paper, we presented our Wildfire Detection and Alert System which can perform highly accurate early detection of wildfire smoke and send immediate alerts to users. Our system is already being used in the USA in a real-world setting to detect and prevent wildfires. We believe that such ML based real-time detection systems are highly important in the fight against climate change as such systems are capable of monitoring data around the clock, detecting events faster than humans, and thus help in arming the first responders with crucial real-time information which allows them to respond faster and help keep people and property safe. Our system is also being used by utility companies to protect transmission lines and other valuable infrastructure assets from wildfire smoke. This ensures that various essential utilities like electricity are provided to people everyday without any interruptions. Therefore, we hope that in the coming years, systems like ours will be used throughout the world to fight against the upcoming threat of more severe wildfires.



\bibliographystyle{unsrtnat}
\bibliography{sources}

\appendix
\section{Appendix}

\begin{figure}[!hbt]
  \centering
  \includegraphics[width=\linewidth]{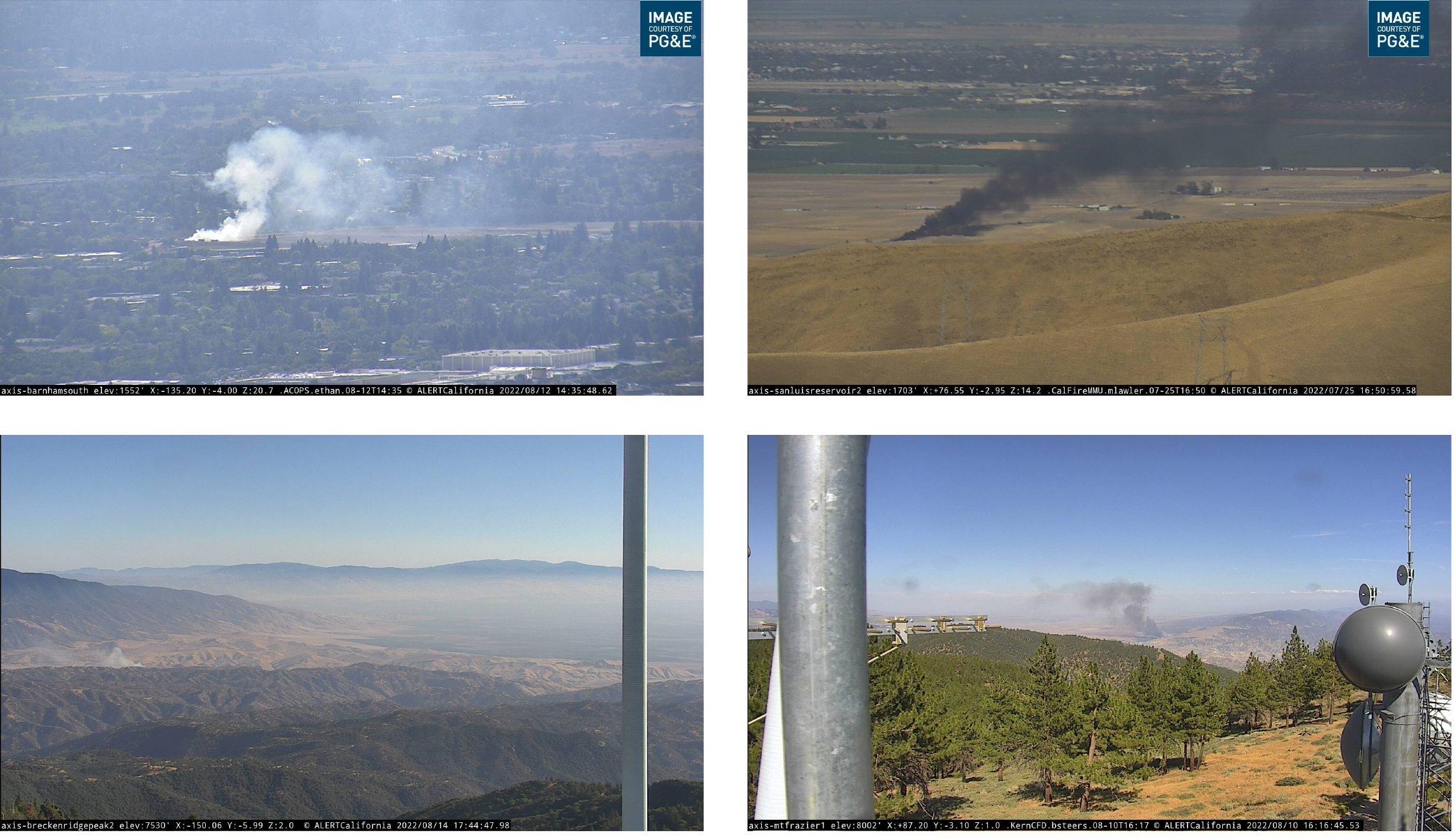}
  \caption{Examples of diverse set of images present in our dataset}
  \label{terr}
\end{figure}

\begin{table}[!hbt]
  \caption{Distribution of the data}
  \label{dataset_dist}
  \centering
  \begin{tabular}{llll}
    \toprule
             & Wildfire     & Non-Wildfire & Total \\
    \midrule
    Train    & 26690  & 19726 & 46416 \\
    Validation     & 11924 & 29210 & 41134 \\
    \bottomrule
  \end{tabular}
\end{table}

\begin{figure}
  \centering
  \includegraphics[width=\linewidth]{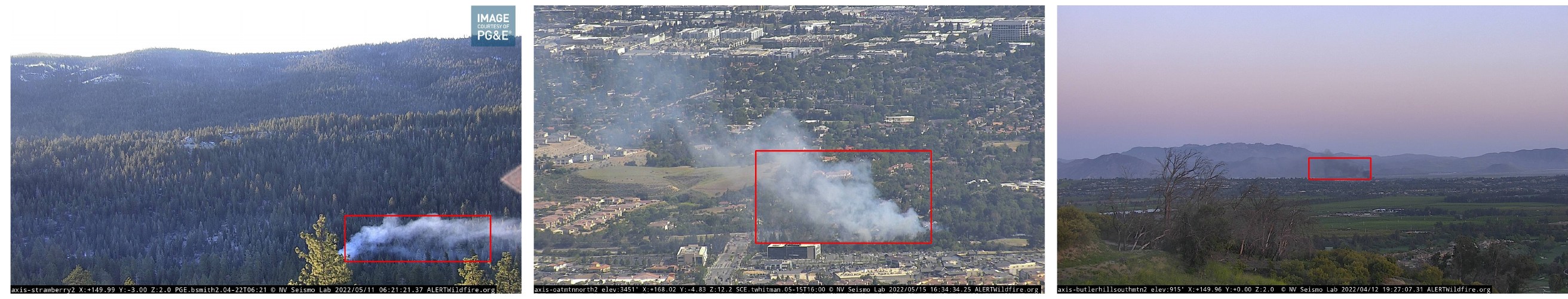}
  \caption{Examples of true positives}
  \label{res_tp}
\end{figure}

\begin{figure}
  \centering
  \includegraphics[width=\linewidth]{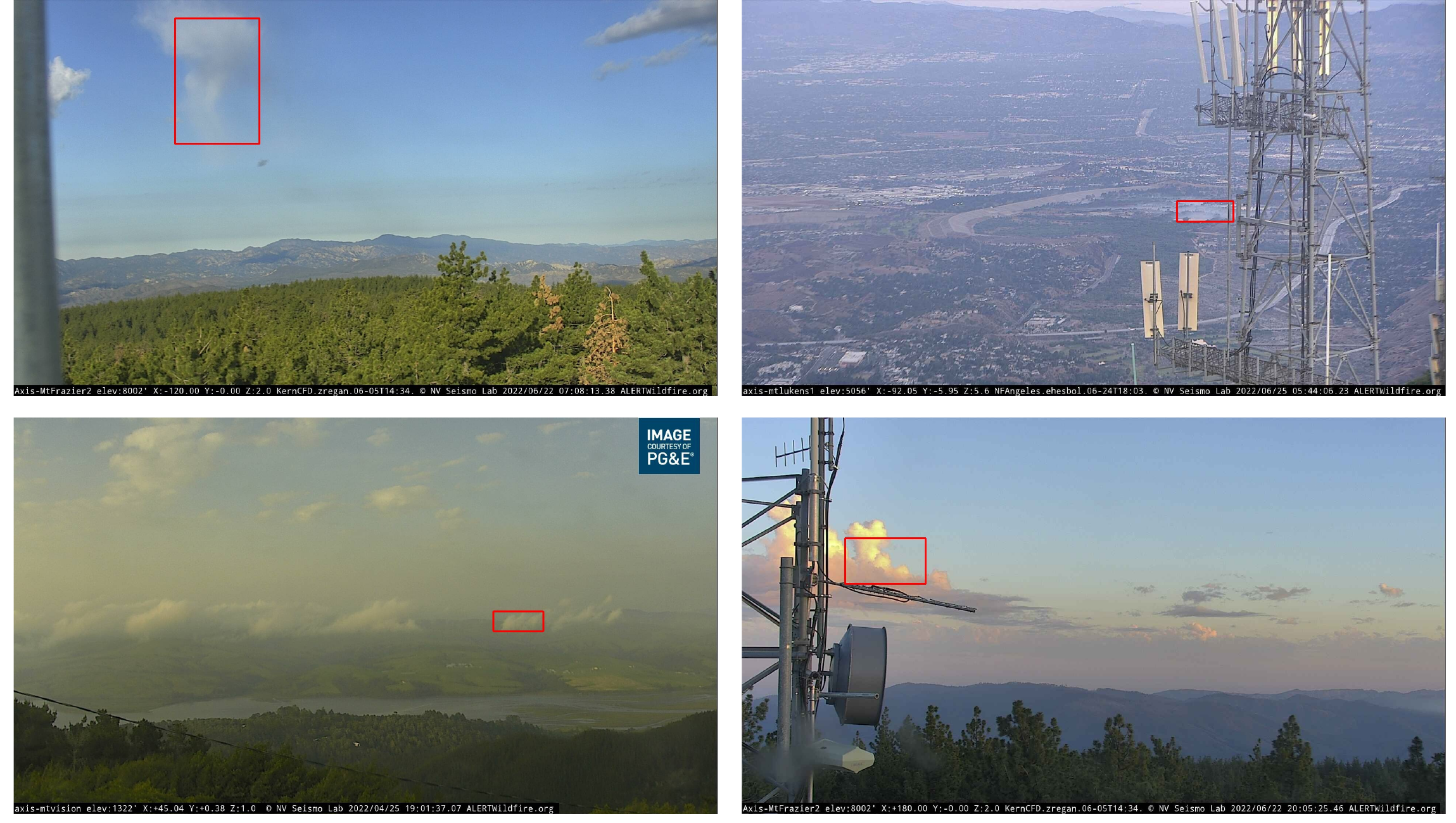}
  \caption{Examples of false positives}
  \label{res_fp}
\end{figure}

\subsection{The Dataset}\label{dataset} 
We collected a total of 90,000 images from Alert Wildfire cameras \cite{alertwild} and labeled them. The images were categorized as wildfire and non-wildfire and then divided into train and validation sets as shown in Table \ref{dataset_dist}. The dataset includes images from a diverse range of terrains as shown in Fig \ref{terr}.
 
\subsection{Detection Results}\label{detRes}
The problem of smoke detection is especially hard in this setting as it can be seen from Fig \ref{terr} and Fig \ref{res_fp} how easy it is to confuse smoke with clouds or fog. However, our model does a pretty good job in general and has low false positives as seen in Table \ref{perf}. Fig \ref{res} and Fig \ref{res_tp} show a few examples of true positives detected by our model. Fig \ref{res_fp} shows a few examples of false positives detected by our model. 

\subsection{Challenges and Future Work}\label{cfwork}
Detecting wildfire smoke in such diverse scenarios is a challenging task. As seen from Fig \ref{res_fp}, it is possible to mistake  clouds, fog, and smoke from industrial processes (e.g. smoke coming from chimneys) as wildfire smoke. 

Currently, the way we tackle false positives is by having a human in the loop. A smoke detection alert is sent to a human operator and only when the human confirms the alert, it gets sent to other users like first responders (Fig \ref{system}). 

Just like false positives, there could also be false negatives, i.e, the system fails to detect the wildfire smoke. However, we found that the system eventually manages to detect the smoke, i.e, the system performs late detection once the smoke gets bigger. So even if the system fails to detect the smoke at time step t, it manages to detect it in most cases at later time steps. This was validated from our evaluation study (Table \ref{eval}) where our system was successfully able to detect all wildfire events in 5+ minutes from the start of the smoke. 

However, we acknowledge that we need to develop better ways to deal with false positives and false negatives and make our model more robust. Therefore, in the future, we plan to gather more data which deals with these scenarios better and train our model further to reduce the rate of false positives/negatives. We also plan to experiment with different ML techniques which could help improve our model performance and incorporate them in future versions of our system. 
 
\end{document}